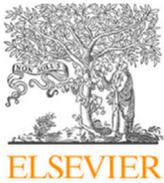
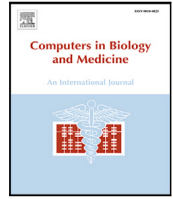

# A store-and-forward cloud-based telemonitoring system for automatic assessing dysarthria evolution in neurological diseases from video-recording analysis☆

Lucia Migliorelli [a,b,*,1], Daniele Berardini [a], Kevin Cela [a,b], Michela Coccia [c], Laura Villani [d], Emanuele Frontoni [e,b,g], Sara Moccia [f]

[a] *Department of Information Engineering, Univerisità Politecnica Delle Marche, Via Brecce Bianche 12, Ancona, 60121, Italy*
[b] *AIDAPT S.r.l, Via Brecce Bianche 12, Ancona, 60121, Italy*
[c] *Centro Clinico NeuroMuscular Omnicentre (NeMO), Fondazione Serena Onlus, Via Conca 71, Torrette (Ancona), 60126, Italy*
[d] *Department of Neuroscience, Neurorehabilitation Clinic, Azienda Ospedaliero-Universitaria delle Marche, Via Conca 71, Torrette (Ancona), 60126, Italy*
[e] *Department of Political Sciences, Communication, and International Relations, Università Degli Studi di Macerata, Via Giovanni Mario Crescimbeni 30, Macerata, 62100, Italy*
[f] *The BioRobotics Institute, Department of Excellence in Robotics and AI, Scuola Superiore Sant'Anna, Piazza Martiri della Libertà, 33, Pisa, 56127, Italy*
[g] *NeMO Lab, Piazza dell'Ospedale Maggiore, Milano, 20162, Italy*

ABSTRACT

**Background and objectives:** Patients suffering from neurological diseases may develop dysarthria, a motor speech disorder affecting the execution of speech. Close and quantitative monitoring of dysarthria evolution is crucial for enabling clinicians to promptly implement patients' management strategies and maximizing effectiveness and efficiency of communication functions in term of restoring, compensating or adjusting. In the clinical assessment of orofacial structures and functions, at rest condition or during speech and non-speech movements, a qualitative evaluation is usually performed, throughout visual observation.
**Methods:** To overcome limitations posed by qualitative assessments, this work presents a store-and-forward self-service telemonitoring system that integrates, within its cloud architecture, a convolutional neural network (CNN) for analyzing video recordings acquired by individuals with dysarthria. This architecture – called facial landmark Mask RCNN – aims at locating facial landmarks as a prior for assessing the orofacial functions related to speech and examining dysarthria evolution in neurological diseases.
**Results:** When tested on the Toronto NeuroFace dataset, a publicly available annotated dataset of video recordings from patients with amyotrophic lateral sclerosis (ALS) and stroke, the proposed CNN achieved a normalized mean error equal to 1.79 on localizing the facial landmarks. We also tested our system in a real-life scenario on 11 bulbar-onset ALS subjects, obtaining promising outcomes in terms of facial landmark position estimation.
**Discussion and conclusions:** This preliminary study represents a relevant step towards the use of remote tools to support clinicians in monitoring the evolution of dysarthria.

## 1. Introduction

Dysarthria is a collective name for a group of neurological speech disorders that reflects abnormalities in the strength, speed, range, steadiness, tone, or accuracy of movements required for the breathing, phonatory, resonatory, articulator, or prosodic aspects of speech production. The responsible neuropathophysiologic disturbances of control or execution are due to abnormalities that often include weakness, spasticity, incoordination, involuntary movements, or excessive or variable muscle tone [1,2]. Neurodegenerative diseases (like amyotrophic lateral sclerosis (ALS) and Parkinson's disease), inflammatory conditions (like multiple sclerosis) and vascular pathologies (like stroke) are the leading causes for dysarthria onset [3].

Offering continuity of care to patients suffering from dysarthria is relevant to avoid the development of extreme social hardship conditions that could directly affect the patients and their caregivers [4]. In fact, the impaired communication ability caused by the onset and evolution of dysarthria undermines an individual's possibility of keeping and extending social contacts, which may eventually undermine his/her overall well-being [5]. Given the premises, the clinical literature recognizes the assessment of dysarthria evolution as a useful tool to: (i) monitor the disease trajectory and properly stage the management of restoring or compensatory strategies as to maximize the effectiveness and efficiency of communication functions, (ii) have new outcome measures for clinical trials, and (iii) detect any correlations with other bulbar signs [6–9].

To search for new quantitative outcome measures to assess dysarthria progress, different approaches were proposed. These mainly monitor the speech and vocal features of dysarthric subjects, both in home and hospital scenarios [7,10–14]. However, as stated in [15,16], also the assessment of orofacial motor functions related to speech (or motor speech assessment) should be considered to: (i) detect subtle improvements or worsening in patients' conditions (especially for those who suffer from ALS, spinal muscular atrophy (SMA), facial palsy and stroke); (ii) evaluate pharmacological and non-pharmacological treatment progress and (iii) improve the staging of the rehabilitative strategies management and pursue an augmentative communication (AAC) assessment [1].

Despite the proven significance, the motor speech assessment mainly relies upon visual observation by clinicians combined with the drafting of rating scales, such as the Robertson dysarthria profile [15,17]. This may lead to inaccurate outcomes as rating-scale descriptors are unlikely to exactly fit a patient's performance [18]. The outcomes are collected during outpatient assessment thus they may be influenced by patients' emotional and physical status at the time of examination [19]. Additionally, the qualitative examination lacks reproducibility and may be prone to intra- and inter-clinicians' judgments variability, since it is influenced by the clinicians' expertise [20].

A possible solution to attenuate the issues caused by the subjectivity of current motor speech assessment has been proposed by the authors in [21]. They employ electromyographic sensors placed over the facial surface and inside the oral cavity. Despite the promising results, the electromyographic acquisition system is complex and the overall acquisition process can be deeply invasive to the patient and not usable in home environment [2].

To objectively and non-intrusively examine orofacial impairments in patients suffering from neurological diseases, the work in [22] proposes a deep-learning (DL) methodology for assessing facial alignment from RGB videos of patients with ALS and stroke. The authors in [22] released their dataset – namely the Toronto NeuroFace dataset – which is a collection of RGB frames with the associated annotations of 68-facial landmarks position. The Toronto NeuroFace is the first annotated dataset in the field and the authors release it to foster the scientific community to propose advanced support methodologies for motor speech examination in patients with ALS and stroke as a way to quantitatively stage dysarthria evolution.

Following [22], our work proposes an innovative cloud-based store-and-forward – i.e., asynchronous – telemonitoring system called Homely Care, aimed at supporting clinicians in dysarthria assessment. Homely Care has two main components: a web application and a scalable cloud architecture which relies on the Amazon Web Services (AWS) cloud platform. The former guides a subject in performing specific assessment tasks from the Robertson dysarthria profile (e.g., keeping lips protrusion for 5 s), the latter automatically captures the video recordings of the subject fulfilling the tasks and sends it to a proposed end-to-end convolutional neural network (CNN). The CNN, which we called facial landmark Mask RCNN, processes the RGB frames from the recordings and outputs the position of the 68 facial landmarks. These are used as prior for assessing new outcome measures of motor speech assessment which may give relevant insights to monitor the trajectory of dysarthria evolution and identify the time to implement rehabilitative treatment or AAC strategies.

The rest of the paper is organized as follows: Section 2 shows the relevant state-of-the-art contributions in the fields of dysarthria assessment, store-and-forward telemonitoring of neurological diseases via consumer devices, and facial-landmark detection. Section 3 presents the implemented methodology (from the CNN description to the cloud-architecture presentation) while technical details for enabling fair comparisons are shown in Section 4. The results achieved are presented in Section 5 and discussed in Section 6. Section 7 concludes the work and proposes the future developments.

## 2. State of the art

### 2.1. Dysarthria assessment

Dysarthria assessments, both in Italy and abroad, relies on clinical-rating scales based on the observation of speech muscles (i) at rest, such as the face muscles at rest (lips, jaw), (ii) during speech activities (henceforth referred to as "speech tasks"), such as spontaneous speech, vowel prolongation, and reading, and (iii) during non-speech activities (henceforth referred to as "motor tasks"), such as sustained posture of lips (e.g., spontaneous smiling, volitional smiling, lip rounding, lip retraction) and jaw (e.g., mouth opening), and diadochokinetic tasks (e.g., jaw alternating motion rates) [1,23]. As introduced in Section 1, such an assessment based on the use of clinical rating scales has, above all, the main limitations of: suffering from ceiling effect – namely it is hard to perceive fine changes in patients' performance at disease's early stage – and hampering the possibility of research during clinical trials by relying on qualitative ratings [24].

For alleviating the above-mentioned limitations, a number of supportive systems to assess dysarthria via patient's speech or vocal-performance analysis have been proposed in literature. Examples include [12,25,26] which investigate the feasibility of machine learning (ML) methods for the analysis of audio-data collected in hospital and home scenarios. In [27,28], a telemonitoring-based application is introduced to automatically assess the evolution in the intelligibility of the speech of dysarthric patients. Also in [13] the authors proposed a methodology for estimating the number of repeated syllables during a diadochokinetic task in subjects with dysarthria.

All the proposed applications, which can potentially be used in a home environment but are mainly tested in controlled scenarios, achieved promising results. However, the state-of-the-art studies offer complementary characterizations of dysarthria as they focus on specific aspect of dysarthria assessment (e.g., articulatory slowness, intelligibility) and do not envisage – unlike the system we propose – a possible extension to other features which may offer clinicians a comprehensive assistive assessment tool. This is particularly relevant for individuals with ALS, SMA, stroke, and facial palsy in whom the motor speech assessment is crucial to capture improvements or worsening and assess possible correlations with other bulbar signs (e.g., dysphagia and respiratory aspects) [17].

### 2.2. Facial landmark detection: from methods to the challenges of self-service telemonitoring

The facial-landmark detection from video recordings is a field of research with numerous clinically-oriented applications ranging from human expressions recognition [29,30] to fatigue detection, [31] and facial-palsy rating [32].

In [33], the authors propose a ML-based methodology for assessing the position of facial landmarks to detect facial palsy. In spite of the promising results, using standard ML instead of DL may lead to generalization problems on videos captured in non-optimal environments [34] (e.g., with variable lighting and background) and may require a hand-crafted features extraction step that makes the approach unsuitable for the actual clinical practice [35,36].

To solve these issues, the work in [37,38] proposes a cascading CNNs-methodology. Their framework, firstly, acts by roughly locating the positions of the landmarks and, then, refines them via a regression sub-network. This methodology, as highlighted in [39], suffers when dealing with poor-lighting conditions that may occur in the home environment, occluded face-parts which may be due to the presence of clothing covering face portions, and challenging landmark positions, which, in the case of neurological patients, may be due to the presence of severe orofacial impairments. Furthermore, the use of two subsequent networks could be inefficient and overtaken by multi-task networks [40].

Recent literature contributions [39,41,42] tackle the challenges posed by real-life scenarios, benefiting of end-to-end CNN-based frameworks which are originally designed to estimate people's pose. Inspired by these research hypotheses, our methodology refits the pipeline in [43] to locate facial landmarks from ALS and stroke patients from the Toronto Neuroface dataset [22].

The designed CNN was then deployed on the Homely Care cloud-based system for being further tested on acquisitions carried out by bulbar-onset ALS patients in the home environment, as to face challenges such as the usage of cameras integrated in consumer devices (e.g., smartphone, tablet, PC), the autonomous management of the acquisitions, varied background and illumination conditions.

*2.3. Store-and-forward telemonitoring of subjects suffering from neurological disorders by consumer devices*

From 2020, telemedicine made significant advancements to uphold the delivery of regular care for neurological patients, particularly those with neuromuscular conditions. Numerous strategies were suggested for various diseases. However, telemedicine emerged as a valuable tool in managing neurological disorders during the pandemic era. Indeed, these tools played a pivotal role in overcoming geographical barriers and pandemic-related challenges by enabling the maintenance of the patient-clinician relationship while minimizing risks associated with in-person evaluations [44,45]. In addition to its role in supporting in-person assessments by offering information on patients in the home environment, telemedicine can be further explored in the development of telehealth applications for continuous remote patient monitoring or self-service telemonitoring. By utilizing devices or apps, patients may provide clinicians with daily updates on their health status. This information proves to be relevant in monitoring general indicators (e.g., blood pressure, cardiac rhythm, and oxygen saturation) but also ambulatory performance like speed and endurance in speaking [45]. Specifically, [46] leverages the myParkinsoncoach application to examine patients through the use of questionnaires. In their work, the authors proved that using Likert scale questions to telemonitor patients decreases the need for outpatient healthcare services among individuals with Parkinson's Disease. The myParkinsoncoach system can be used from any platform – and thus device – and is designed to provide the clinician with an intuitive interface to view trends generated by patient responses. However, the evaluation of the patient's responses alone is not enough especially to assess dysarthria, where a comprehensive solution is crucial.

Following this consideration, [47] presents Apkinson, a mobile application for the Android operating system. This is designed for speech and walking assessment of subjects with Parkinson. Apkinson takes advantage of the device built-in sensors to acquire data that are processed by signal-processing algorithms; it is designed to be usable in the home environment in a store-and-forward fashion but is tested in a controlled scenario. Despite the multimodal nature of the solution, the application is designed to be used only by smartphones running the Android operating system. Furthermore, the authors state that the system is not meant to be scalable and therefore cannot handle too many requests nor can it be easily integrated with other assessments. Similarly the work in [48] proposes a system for telemonitoring subjects with Parkinson's disease. Their system uses the accelerometer and microphone of the smartphone to record multimodal data that are analyzed by machine learning algorithms. As in [47] the proposed system is envisioned for use in the home environment but is actually tested in a controlled environment while few references are made to the cloud platform that is supposed to manage the entire flow.

Currently the preprint paper of Neumann et al. [49] proposes the closest solution with respect to ours. Indeed, with NEMSI the authors present an internet-based multimodal dialogue system designed to gather necessary evidence for the identification or ongoing monitoring of patients with ALS. To this end, the authors highlight how for ALS – unlike Parkinson's disease in the previously mentioned approaches – the assessment of orofacial functions related to speech is crucial to detect subtle changes in patients' performance as well as disease trajectory. In line with this research, Homely Care automatically performs the motor speech assessment of a dysarthric subject through video recordings acquired with a web application in communication with a cloud architecture. Given the inherent modularity and scalability, our proposed system – which in this work automatically analyzes video data via a proposed facial landmark mask RCNN – may, by design, already safely collect other data (e.g., audio) to provide a comprehensive assessment of the disorder. Moreover, the application can be used by any personal device equipped with a camera and microphone with no limitations on the operating system while the cloud architecture is designed and implemented to easily and safely manage data from multiple connected users.

**3. Methods**

*3.1. Facial landmark mask RCNN*

Our pipeline for facial-landmark detection is shown in Fig. 1 and was inspired by Mask RCNN [43], which was originally designed to predict the pose of the human body in two-dimensional space. Here, Mask RCNN was modified to detect the facial landmarks.

As for Mask-RCNN, the proposed pipeline is made up of two stages. The first stage is used to extract the regions of interest (RoIs), while the second stage is used to further refine the localization of the face and to detect facial landmarks. To this end, the proposed CNN has 3 main branches: one for classification, another for bounding-box regression and the last for facial-landmark position estimation. In the first stage, the RGB-input image is fed into a backbone-CNN. The backbone is a ResNet50 (i.e., ResNet with 50 convolutional layers), coupled with a feature-pyramid network (FPN), acting as feature extractor [50].

These output feature maps are fed into a region proposal network (RPN), which generates RoIs. Then, each region proposal is sent to the RoI Align layer which generates a small fixed-size feature map from each RoI. Warped features in output from the RoI Align are then fed into fully connected layers of the second stage. These layers output the bounding-box coordinates and the label category (i.e., "face") with the relative prediction-confidence in output from the soft-max layer of the classification branch. It is worth noting that this stage of bounding-box regression is crucial when deploying the pipeline in scenarios (e.g., hospital wards) where empty backgrounds may not be guaranteed. In parallel with the bounding box coordinate regression and class assignment, warped features are also fed into the mask branch which is a fully convolutional network properly adapted for the facial-landmark detection task. Particularly this branch was modified, compared with the original implementation, for the task of interest by adding two subsequent strided ($s = 2$)-transposed convolutions (with kernel sizes $3 \times 3$ and $4 \times 4$, respectively). The last convolutional layer outputs

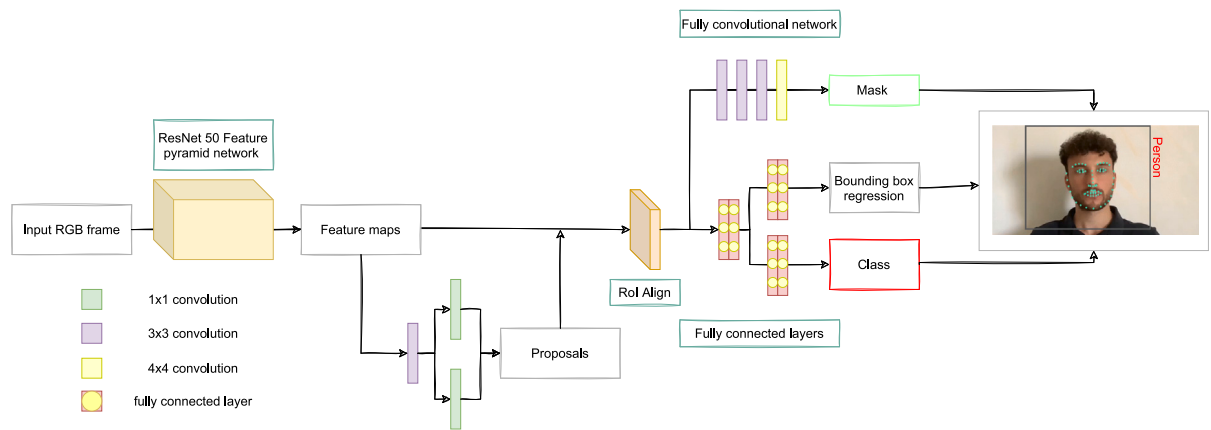

**Fig. 1.** Facial landmark mask RCNN.

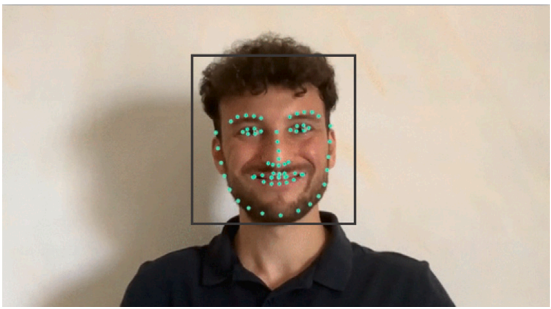

**Fig. 2.** Sample image with the facial landmarks (green dots) and the bounding box for face localization (black square).

68 binary masks, one for each landmark. The use of two additional transposed convolution layers allows the resolution of the output map to be extended facilitating the network to locate the 68 face landmarks.

The following paragraphs outline the datasets used to train and validate the proposed facial landmark mask RCNN.

*3.1.1. The 300 videos in-the-wild dataset*

To train our facial landmark mask RCNN the fine-tuning technique was adopted. To this end, the network was pre-trained on a subset of the 300 Videos In-the-Wild Challenge (300-VW) dataset [51].² This is a publicly available dataset of 114 videos showing one person each. Each video has a duration of 1 min and is recorded in 30 frames per second (FPS). The dataset has three logical categories of videos, with increasing challenges for each category. The first category contains videos of people acquired in well-lit conditions in various head poses, without occlusions. The second category contains videos of people acquired with variable light conditions (different illuminations, dark rooms, overexposed shots, etc.) with minimal occlusions. The third category contains videos of people acquired in unconstrained conditions, including highly variable illumination conditions, occlusions, make-up, expressions, and head poses. For our purposes, 1 out of 5 frames of each video from the original 300-VW was selected with a training, validation and testing set split of ∼70%, ∼20% and ∼10%, respectively. The weights resulting from the pre-training on such 300-VW subset were then used to initialize the network.

*3.1.2. The Toronto NeuroFace dataset*

The fine-tuning was conducted on the Toronto NeuroFace dataset [22], a collection of RGB video recordings from ALS patients (11

---

² https://ibug.doc.ic.ac.uk/resources/300-VW/

subjects: 4 males, 7 females), stroke patients (14 subjects: 10 males, 4 females), and healthy subjects (11 subjects: 7 males, 4 females) of similar age.

The video recordings were acquired in a controlled environment with optimal lighting conditions with an Intel® RealSense camera placed 30–60 cm away from the subject's face. Each subject was asked to perform specific speech and oral motor assessment tasks, such as keeping maximum mouth opening and lips stretching, and repeating the syllables /pa-ta-ka/ or /pa/.

Frames were extracted from each video recording as to maximize intra-subject variability. For instance, for repetitive motor tasks only three frames per-repetition were considered, one showing the gesture beginning, one for its peak, and one between them. After frames extraction, manual annotation of the 68 facial landmarks and the face bounding box (Fig. 2) was performed for 3306 frames, of which: 1015 frames with healthy subjects, 920 frames with ALS individuals, and 1371 with stroke individuals.

For the purpose of this work, the Toronto NeuroFace dataset was split in training, testing and validation set keeping frames from 32 subjects to train and validate the CNN, and frames from 4 gender-balanced subjects (of which 2 with ALS and 2 with stroke) to test it.

*3.2. The homely care cloud-based store-and-forward telemonitoring system*

*3.2.1. Web application*

To enable multimedia-data acquisition in the home environment, we designed and implemented the Homely Care application. This is a web application (therefore, from now on, we will refer to it as sayweb app) developed in Typescript using the library ReactJS. Subjects involved in the study used it on their personal smartphones, tablets or PCs. The tasks in the web app were a subset of those in the Robertson dysarthria profile [15], and were selected by speech language pathologist as those that are functional to assess the disorder in individuals with ALS. Among the selected tasks were: lips stretching, lips protrusion, maximum mouth opening, mouth opening and closing, mouth protrusion and stretching. The web app provides instructions on how to start a recording as well as tutorials by a speech language pathologist to support subjects in executing each task correctly.

Participants were registered by their clinicians in the system before conducting acquisitions for the first time. After entering the relevant data for account registration, each individual used the web app autonomously. After logging in, the patient was able to visualize the assessment sessions to be performed (left side of Fig. 3). Specifically, by design, the clinician dictated a timetable of acquisitions and the patient could see on a daily basis when he or she had to perform a new session.

When starting an acquisition session, the web app sequentially shows the patient the tasks to be performed, each marked by a title,

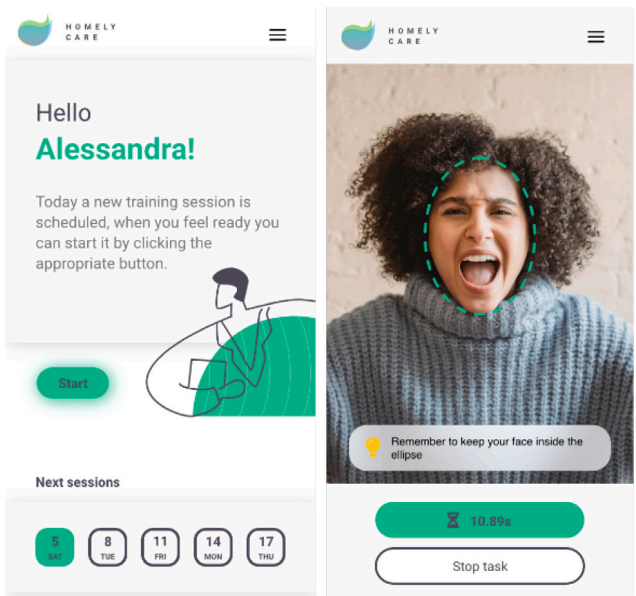

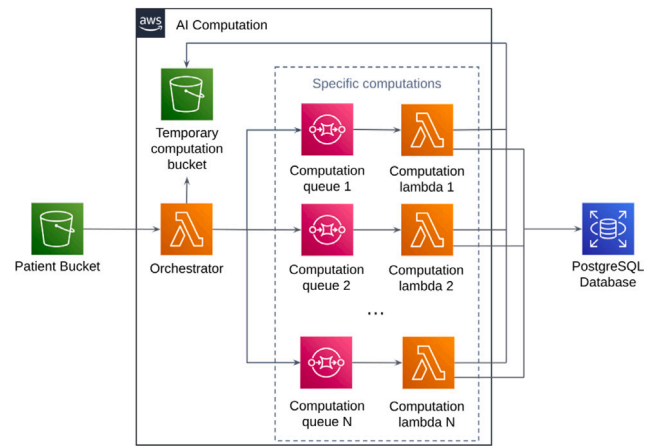

**Fig. 3.** Screenshots of the Homely Care web application (web app), enabling users to perform the Robertson-dysarthria-profile tasks for assessing orofacial muscles. The frame on the left part shows the starting page of the application. From this page the user can both visualize the calendar of acquisition sessions and start a new one. The right side the frame shows the acquisition screen. The user has the option to press play when they are ready to perform the test, and stop the acquisition at any time (except for tasks that have a fixed duration). During the execution of the task, the application automatically records the video stream and guides the user in the correct positioning of the face through an ellipse that appears on the screen. Users can start the acquisition when the ellipse is colored green (i.e, correct positioning of the user with respect to the camera's field of view).

**Fig. 4.** Diagram representing the cloud architecture for the distribution and processing of multimedia data via deep-learning methodologies. Icons are from official Amazon Web Services (AWS) suites.

a description, and a maximum or predefined duration. The web app films the patient during each task, only after a countdown to let the individual adequately prepare. The web app also allows the user to eventually discard and repeat the acquisition if adverse events occurs, such as the sudden appearance of another subject in the camera field of view. At the end of each task, a video file related to the task is stored. Upon completion of all the tasks, all the video data are compressed into a single file in .zip format. This .zip file is sent to the storage service for being subsequently processed by our facial landmark mask RCNN. If the file is successfully uploaded, the session is completed and the .zip file is discarded from local storage. If, on the other hand, the upload is unsuccessful, the web app put the file to be uploaded in a queue and, when possible, proceeds with its upload again.

The web app does not impose stringent constraints (e.g., homogeneous and clear background) on the acquisitions, rather, it suggests the patient to record in a bright environment, in order for the facial landmark Mask RCNN to work in optimal conditions. In addition, before starting the acquisition, an ellipse appears in overlay on the screen to guide a user in the correct positioning of the face with respect to the camera field of view. The acquisition does not start until the subject's face is not correctly positioned within the ellipse (right side of Fig. 3).

### 3.2.2. Cloud architecture

The cloud architecture, built using AWS's services,[3] was designed to meet the following essential functional requirement:

- The system must manage the .zip files of the sessions, save them in a storage, and guarantee characteristics such as persistence, integrity, and even the possibility of encrypting the data, in order to ensure the protection of personal data, in compliance with the General Data Privacy Regulation[4];
- The files within the storage must be available for consultation by clinicians, to check, for example, the evolution of the patient's condition and the correctness of the acquisitions;
- The system must expose an interface that allows the web app to upload the .zip file with the data to be then processed by the CNN. The scheme for distributing and processing a file must be easily extensible, allowing the addition of new tasks and algorithms for data analysis and, thus, the system expansion;
- The system must have a database capable of storing the information of interest and managing requests and transactions coming from the clients;
- The system shall present a backend to allow proper communication with the web app, providing Create, Read, Update and Delete (CRUD) functionality on web app-relevant data;
- The system must correctly manage the registration and login of users.

Given the requirements, lambda functions were considered for handling data processing via the proposed facial landmark mask RCNN. Concerning the automatic distribution of data from each of the task, we identified a solution based (i) on the use of queues (see "computation queue" in Fig. 4), defined via the Amazon Simple Queue System (SQS) service, and (ii) on the design of a structure consisting of a main lambda function or "orchestrator", and a set of lambda functions, namely "consumers" or "computation lambda" (See Fig. 4), connected to the queues. By using queues associated with the lambda functions, we guaranteed system scalability as the resources instantiated for the operation of the lambda function are proportional to the messages remaining on the queue yet to be processed.[5] This feature is particularly relevant when dealing with DL algorithms for data processing. Indeed, such processing may not be immediate for the individual file and having a system able to offer good scalability as the amount of files to be processed increases is a crucial need. Fig. 4 shows a diagram of the previously stated design. Each consumer implements a unique multimedia-data analysis algorithm. In this case, the algorithm is the facial landmark mask RCNN, which outputs the 68 facial landmark positions from the RGB input frames. However, the system is designed to potentially integrate other assessment tasks extracted from Robertson dysarthria profile and, consequently, other algorithms for analyzing the newly collected data. Each of these consumers is connected to an SQS queue, which is used for the collection of messages indicating the location of the media files to be processed within a special bucket, deployed as temporary storage for file processing.

---

[3] https://aws.amazon.com/

[4] https://commission.europa.eu/law/law-topic/data-protection/data-protection-eu_en

[5] https://docs.aws.amazon.com/lambda/latest/dg/with-sqs.html

In particular, each consumer performs a *polling* of the associated queue to receive a given number of messages (batch), which will be associated to the processing of the file. Depending on the number of messages sent in the queue, the consumer can automatically scale. Following algorithmic processing, the consumer saves the result within the system database for enabling the clinicians to access and view the data. After this processing, the file is deleted from the temporary bucket.

The orchestrator, on the other hand, is the lambda function that actually performs the distribution of data to one or more consumers. The orchestrator monitors the patient's bucket to detect the loading of new files associated with each session. Therefore, when the .zip file with the patient's data of the entire session is loaded by the web app, the orchestrator (i) adds a new session in the database, with a timestamp (in ISO format 8601 format) equal to the time of the data receipt, (ii) decompresses the received file onto the temporary storage bucket, (iii) selects specific consumers for data processing, and (iv) sends messages to each of the queues associated with the consumers aimed toprocess the file. This structure, in addition to allowing automatic and scalable computation, has the advantage of being easily extensible: to add a new analysis algorithm to the system (for example, for audio data processing), it is sufficient to add a new consumer and the associated queue, and modify the database to add the new index, binding it to the lambda function that will process it.

## 4. Experimentation

### 4.1. Training and deployment settings

The facial landmark Mask RCNN was pre-trained on the 300-VW dataset for 100000 iterations using stochastic gradient descent (SGD) as optimizer. The initial learning rate was set to 0.005, with a learning-rate decay of 0.5 applied every 25% of the total number of iterations. The resulting network weights were used as a starting point for the fine tuning on the Toronto NeuroFace dataset. The fine tuning was carried out in 28000 iterations. The initial learning rate was set to 0.001 with a learning-rate decay of 0.5 every 15000 and 20000 iterations, respectively. The number of maximum subjects to detect per image was limited to 1. The confidence threshold for the bounding box was set to 0.75. All these training settings come from an extensive grid-search to find the best combination of loss, optimizer, learning rate scheduling, and iterations.

Online data augmentation was used to increase the size and variability of the Toronto NeuroFace dataset. Random brightness (with brightness factor ranging from 0.8 to 1.2) and flipping (with a probability equal to 0.5) were considered.

To perform the computation on cloud, the AWS lambda functions leverage an arm64 architecture equipped with AWS Graviton2 processor and 4 GB of memory, resulting in a CPU processing time of roughly 10 min for each acquired task. It is worth noting that, as the clinical assessment does not require real-time outcomes, we based our system on the store-and-forward paradigm. Additionally, as per our clinical partners' agreement, the web app should refrain from providing patients with quantitative performance feedback regarding task execution. These factors allowed us to scale computational cloud resources to a minimum with significant cost savings.

### 4.2. Ablation studies

The performance of our facial landmark mask RCNN was compared against the performance of (i) its akin, namely N-FLMask, trained from scratch (i.e, without pre-training on the 300-VW dataset) and (ii) the original Mask RCNN (i.e., without changing the regression branch for landmark-position estimation) trained from scratch (i.e, without pre-training on the 300-VW dataset), namely N-MaskRCNN.

Then, the performance of our network was compared against the performance of the 300VW-FLMask (i.e., the facial landmark mask RCNN trained on the 300-VW dataset only, without fine-tuning on the Toronto Neuroface dataset).

An overview of the ablation studies is shown in Table 1. For each CNN the training settings are those described in Section 4.1. All the networks share the same training, validation and test set split for fair comparisons.

### 4.3. Evaluation metrics

To assess the performance of the tested CNNs, the Normalized Mean Error ($NME_k$) was computed across all the test images as follows:

$$NME_k = \left[\frac{1}{N_L}\sum_{i=1}^{N_L}\frac{\sqrt{(x_i - xp_i)^2 + (y_i - yp_i)^2}}{Diag_{bbox}}\right] \cdot 100 \quad (1)$$

where $Diag_{bbox}$ is the length of the bounding box diagonal, $N_L$ identifies the number of landmarks, $(x_i, y_i)$ are the ground-truth coordinates and $(xp_i, yp_i)$ are the predicted landmark coordinates. $NME_k$ was assessed for the totality of 68 landmarks ($NME_{68}$), for the 17 chin landmarks ($NME_{chin}$), for the 10 eyebrow landmarks ($NME_{eyebrows}$), for the 9 nose landmarks ($NME_{nose}$), for the 12 eyes landmarks ($NME_{eyes}$) and for the 20 mouth landmarks ($NME_{mouth}$).

### 4.4. Clinical outcomes

The subjects involved in data acquisition via the web app were both males and females with bulbar-onset ALS. They consecutively referred to the facility and started using the application following the acceptance of the study by the Ethics Committee.[6] Study participants used the web app freely and performed the following tasks from the Robertson dysarthria profile: maximum smile, lips stretching and lips protrusion for 5 s, 5-times maximum-mouth opening, opening and closing the mouth for 30 s, protrusion and stretching of the mouth for 30 s. Thus each video lasted at least 30 s, which we empirically sampled at 8 FPS.

We excluded from the study ASL individuals with tracheostomy, percutaneous endoscopic gastrostomy, cognitive impairment, individuals without a caregiver and with concomitant diseases that could interfere with communication skills or could affect life expectancy. Each subject involved in this preliminary study underwent assessment scales as the ALS functional rating scale revised (ALS-FRS-R) which was assessed to determine the overall (*i.e.*, not exclusive to bulbar features) severity of the disease, with a maximum achievable score of 48 [52].

We further performed the Montreal cognitive assessment (MoCA) to evaluate potential cognitive impairments (maximum achievable score = 30) [53] and assessed dysphagia, i.e., the patient's difficulty to swallow foods or liquids, via the dysphagia outcome and severity scale (DOSS) (maximum achievable score = 7) [54]. The characteristics of the bulbar-onset ALS subjects included in our study are summarized in Table 2.

From this qualitative analysis on bulbar-onset ALS subjects we further proposed a prototype measure of orofacial functions related to speech. This measure results from the position of facial landmarks and is computed as the difference between the peak in the performed gesture ($D_p$) and the neutral expression ($D_n$) normalized for the inter-ocular distance ($D_o$). The task considered in this work dealt with keeping maximum lip protrusion for 5 s, thus we named the measure as *LP-index*:

$$LP - index = \frac{D_p - D_n}{D_o} \qquad (2)$$

---

[6] Protocol-ID 118 25/03/2021.

**Table 1**
Ablation studies for validating the proposed CNN for facial-landmark detection in subjects suffering from neurological diseases.

|  | Facial landmark mask RCNN (our) | N-FLMask | 300VW-FLMask | N-Mask |
|---|---|---|---|---|
| Pretraining | 300-VW | x | x | x |
| Fine tuning | Toronto NeuroFace | Toronto NeuroFace | 300-VW | Toronto NeuroFace |

**Table 2**
Characteristics of the 11 bulbar-onset ALS subjects included in the study who consecutively referred to *Azienda Ospedaliero-Universitaria delle Marche* (Ancona, Italy).

| Subject ID | Age | Gender | ALS-FRS-R | MoCA | DOSS |
|---|---|---|---|---|---|
| 1 | 63 | F | 33 | 30 | 3 |
| 2 | 70 | F | 32 | 28 | 5 |
| 3 | 67 | M | 42 | 30 | 5 |
| 4 | 77 | F | 40 | 27 | 3 |
| 5 | 80 | F | 35 | 26 | 4 |
| 6 | 55 | F | 39 | 30 | 3 |
| 7 | 57 | M | 40 | 26 | 5 |
| 8 | 53 | M | 37 | 29 | 5 |
| 9 | 61 | M | 35 | 30 | 6 |
| 10 | 61 | F | 38 | 28 | 5 |
| 11 | 75 | M | 40 | 30 | 5 |

The distances are computed as Euclidean ones in the two-dimensional space.

## 5. Results

The quantitative results of the 4 CNNs (Table 1) on the testing set from the Toronto NeuroFace dataset are shown in Table 3. The performance of the N-FLMask was compared with the performance of the N-Mask to prove the effectiveness of the architectural variation. As can be seen from Table 3, the results of the N-FLMask were significantly better than the results of the N-Mask with a $NME_{68}$ equal to 2.70 against $NME_{68}$ equal to 13.55 achieved by the N-Mask.

To prove the effectiveness of the chosen training protocol (pre-training on 300-VW Dataset and fine-tuning on the Toronto NeuroFace dataset), the performance of our facial landmark mask RCNN was compared with that of the N-FLMask and 300VW-FLMask. As shown in Table 3, our architecture achieved the highest performance with a $NME_{68}$ equal to 1.79 against $NME_{68}$ equal to 2.70 and 3.88 of the N-FLMask and 300VW-FLMask, respectively.

A similar trend in performance is also visible for the $NMEs$ on the other facial regions (i.e., chin, eyebrows, nose, eyes, mouth).

We then deployed our facial landmark Mask RCNN on the Homely Care AWS cloud-based system and we reported in Fig. 5 the qualitative results achieved by our CNN tested on the acquisitions made via the web app by the 11 bulbar-onset ALS individuals consecutively referred to *Azienda Ospedaliero-Universitaria delle Marche*.

## 6. Discussion

Close and quantitative monitoring of dysarthria evolution by assessing the orofacial functions related to speech is crucial to examine the progress of neurological diseases, such as ALS or stroke. However, despite its importance, the monitoring of such muscle functions mainly relies upon visual observation by clinicians combined with the drafting of rating scales. This procedure, besides being qualitative and feasible only during the outpatient assessments, does not allow clinicians to perceive fine changes in patients' performance. To attenuate the issues of subjective assessments, the authors in [21] propose a monitoring approach based on electromyographic sensors placed over the facial surface and inside the oral cavity. Using such sensors may be too invasive for the patient and the examination can only be implemented in a controlled environment. A number of non-invasive systems to telemonitor neurological disorders have been proposed in the literature in the last decade. Some ask questions to the patient, others take advantage of the sensors embedded in the smartphone to evaluate – through signal processing or ML and DL algorithms – the evolution of the disease by assessing biometrics such as tremor, gait and vocal quality [46–48]. However, such approaches have been only partially tested in the home environment, and do not envisage the possibility of scaling the cloud system to integrate new evaluation tasks, or handling multi-platform and multi-user settings. This may hamper the translation of the approaches in the actual assessment practice.

To overcome possible state-of-art limitations and to assess dysarthria in a quantitative non-invasive, and user-friendly fashion, the proposed research presents a self-service store-and-forward telemonitoring system for assessing orofacial functions related to speech. Our Homely Care is based on a scalable and easy-to-integrate cloud platform to safely handle the data, and a CNN aimed to automatically regress the position of facial landmarks from video recordings acquired with a web application usable by any consumer device and operating system.

Thus, our work consists, first, in the design of a CNN for facial landmark detection. Inspired by literature in closer fields [43], we presented our facial landmark mask RCNN. At this stage, we obtained the quantitative results (shown in the Table 3) by testing the CNN on a portion of the Toronto NeuroFace dataset. By the comparison of the results of our facial landmark mask RCNN with the performance of N-FLMask and 300VW-FLMask (i.e., the same network architectures but with different training strategies), it emerged that the choice of pretraining the CNN on the 300-VW dataset and then fine-tuning it on the Toronto NeuroFace dataset was crucial. Indeed, the pre-training on the wider dataset granted the network a higher power of generalization while the fine-tuning allowed to refine the CNN ability to regress facial landmarks from pathological subjects.

The original version of the Mask-RCNN (i.e., N-Mask) got the largest error (i.e., the lowest performance). This suggests that varying the branch for landmark-position regression allowed to recover a higher level of details. It is worth noting that the chin and mouth-related landmarks were the most challenging to detect, and this may depend on the major impact that the progressive-disease has on the muscles of the oral district [22]. In this case, the least flawed of the 4 architectures was our facial landmark mask RCNN. This may be due to the fact that the 300-VW dataset – which was used to pre-train the network – features people grimacing and, therefore, displays orofacial impairments.

After the CNN design, we implemented the cloud-based Homely Care system aimed to remotely monitor dysarthria. The system consists of (i) the web app for video-data acquisition and (ii) the cloud architecture which, besides handling the data acquired via the web app, hosts our facial landmark mask RCNN.

To test the system functionality, the web-app usability, and the CNN performance on data acquired in a real scenario, we let 11 individuals with bulbar-onset ALS use the web app. The obtained results (see Fig. 5) show that our facial landmark mask RCNN is robust to various illumination levels. This is particularly evident from the frames 1, 2, and 3, in which the landmarks are correctly positioned over the subjects' face. Frame 1 shows the patient optimally positioned with respect to a light source while performing a maximum lip-protrusion task. In frame 2, the patient is positioned under an artificial-light source and is performing a lip-stretching task. Instead, in frame 3, the patient is in a neutral position in a dark environment. In the remaining frames we show, on the contrary, a number of inaccuracies. In frames 4 and 8, landmarks are not correctly positioned in the eyebrow area. This may be due to two factors that both subjects share and that are poorly represented in the training datasets: the presence of glasses and the poorly defined eyebrows. In addition, in frame 4, our CNN also makes some localization errors in the area of the mouth. This could be caused by the fact that the patient is performing a task of maximum-mouth opening, thus she is assuming an extreme position during which the labial rim tightens causing the landmarks delimiting the lower and upper lip to come closer together. Frames 5 and 6 show a diadochokinesis task in which the patient was required to perform a sequence of movements for a predetermined time. Specifically, the task involved alternating lip protrusions and stretches. When the patient performs the lip-protrusion task, our facial landmark mask RCNN fails in localizing lips landmarks. This error is highlighted also in literature and may be due to the fact that the patient, in performing the task, naturally develops expression lines that may confuse the CNN which is trained on datasets with few samples such as those shown in the frame [55]. As for frame 7, landmarks are incorrectly localized due to the positioning of the patient in relation to camera field of view. This specific acquisition was conducted with the outdated version of the web app in which there was no ellipse guiding the patient in the correct positioning.

**Table 3**
Results in terms of Normalized Mean Error ($NME$) for each convolutional neural networks (CNNs).

|  | Facial landmark mask RCNN (our) | N-FLMask | 300VW-FLMask | N-Mask |
| --- | --- | --- | --- | --- |
| $NME_{68}$ | **1.79** | 2.70 | 3.88 | 13.55 |
| $NME_{chin}$ | 2.62 | 4.81 | 4.39 | 15.31 |
| $NME_{eyebrows}$ | 0.02 | 0.03 | 0.05 | 0.12 |
| $NME_{nose}$ | 1.55 | 2.08 | 3.60 | 5.61 |
| $NME_{eyes}$ | 1.03 | 0.94 | 3.04 | 5.23 |
| $NME_{mouth}$ | 1.49 | 2.19 | 3.70 | 21.17 |

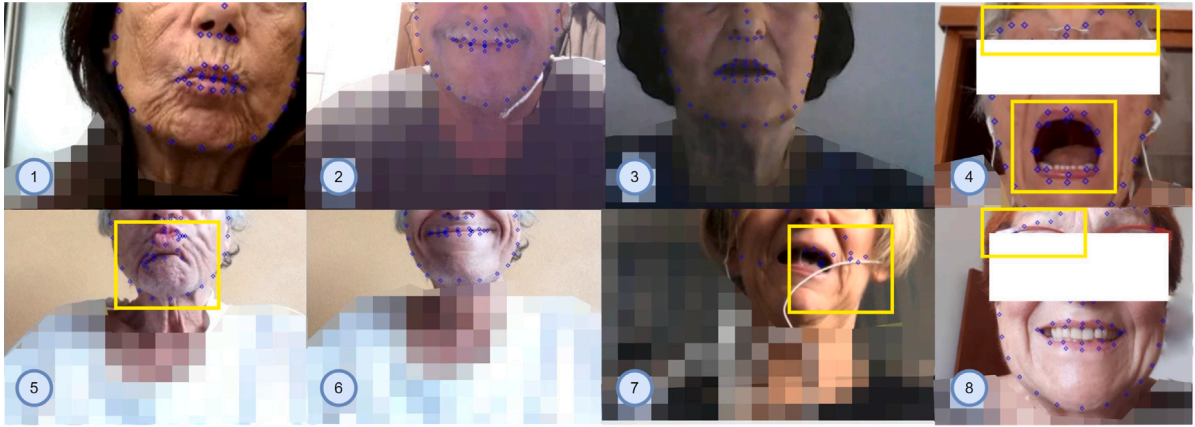

**Fig. 5.** Qualitative outcomes: frames 1, 2, 3 demonstrate how our architecture is robust to various lighting and background levels. Frames 4, 5, 7, 8, show errors made by the facial landmark mask RCNN in estimating the position of the 68 landmarks. Frames 5 and 6 show how, for the same patient, the network makes errors for the specific task of lips-protrusion maintenance. All the errors are highlighted by a yellow box.

A straightforward limitation of the proposed system lies in the limited number of subjects with neurological disease both in the Toronto NeuroFace dataset – that we used to fine-tune our facial landmark mask RCNN – and in our first experimentation conducted on 11 bulbar-onset ALS patients only. However, the purpose of this work was the proposal of a scalable cloud-based store-and-forward telemonitoring system for non-invasive assessment of orofacial speech functions that patients could easily use in the home environment with their personal devices. Future directions of the present research will surely involve more subjects both to expand the dataset to fine-tune the CNN and to quantify with proper surveys the actual usability of the system from both a clinical and patient-side perspective.

## 7. Conclusion and future perspectives

The social and health burden of neurological diseases is set to increase further with the ageing of the population and the epidemiological trends observed over the last 10 years, which confirm exponential increases in incidence and prevalence not only in the elderly population, but also in paediatric and young-adult age groups. The progressive accumulation of disability that characterizes most neurological diseases often prevents adequate therapeutic and care continuity. As a consequence, discontinuity in care may generate serious conditions of social distress involving an increasing number of individuals, families and caregivers [56,57].

To meet the need of guaranteeing care continuity, we proposed Homely Care: a self-service store-and-forward telemonitoring system for evaluating dysarthria evolution via orofacial speech muscle assessment in a quantitative and non-intrusive way. This system integrates with a cloud architecture hosting the proposed facial landmark mask RCNN for facial landmark-position estimation. From the position of facial landmarks, as highlighted in [22], it will be possible to evaluate the impact that diseases – like neurodegenerative ones – have on the muscles involved in vital functions, such as speech articulation and breathing. This will allow to provide decision support to clinicians. In this regard, next improvement of this work will involve the implementation of a dashboard for the clinician aimed at viewing specific indexes to quantify the degree of lips protrusion, lip stretching, maximum mouth opening, facial and mouth symmetry and how these indexes evolve in time. Fig. 6 shows a first prototype of index to quantify the degree of lips protrusion over time.

This research represents a part of a wider project aimed at proposing a comprehensive dysarthria-assessment system that integrates also tasks to quantify other dysarthria features such as speech intelligibility, phonation, respiration, oral diadochokinesis. Especially when dealing with rare diseases, having quantitative and easily accessible data is relevant both for establishing treatment plans specific to patients needs and for improving knowledge about such a disabling disease. Moreover, from patients' side, traveling to the hospital is sometimes tiring, decreases their performance during evaluations phases, and has a real cost. Systems such as the one described in this research are crucial, as they enable to carry out assessments at home, with consumer devices (such as smartphone and PCs) and in a familiar environment, while always keeping patients in virtual contact with their trusted clinician. Particularly, the latter aspect will be soon investigated through structured and semistructured interviews involving both patients and their caregivers.

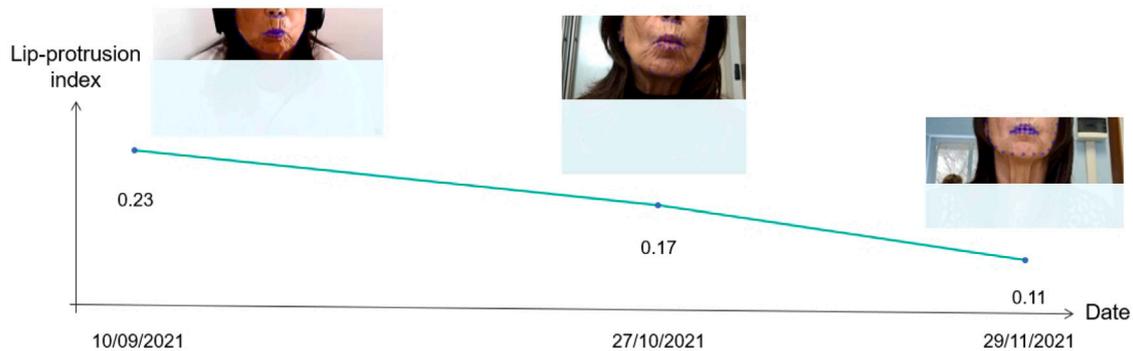

**Fig. 6.** Example of a visualization interface for the clinician. There we show how the prototype of lip-protrusion index varies over time. Images are masked for privacy and commercial logos reasons.

**CRediT authorship contribution statement**

**Lucia Migliorelli:** Conceptualization, Methodology, Writing, Software. **Daniele Berardini:** Writing, Software. **Kevin Cela:** Writing, Software. **Michela Coccia:** Writing, Clinical validation. **Laura Villani:** Writing, Clinical validation. **Emanuele Frontoni:** Writing – original draft. **Sara Moccia:** Writing – original draft, Conceptualization.

**Declaration of competing interest**

The authors declare that they have no known competing financial interests or personal relationships that could have appeared to influence the work reported in this paper.